\documentclass[dvipsnames]{article}

\usepackage[final]{lychee}
\usepackage[utf8]{inputenc} 
\usepackage[T1]{fontenc}    
\usepackage{url}            
\usepackage{booktabs}       
\usepackage{amsfonts}       
\usepackage{nicefrac}       
\usepackage{booktabs} 
\usepackage{multirow}
\usepackage{wrapfig}
\usepackage{amssymb}
\usepackage{epigraph}
\usepackage{makecell}
\usepackage{listings}
\usepackage{subcaption}
\usepackage[labelfont=bf]{caption} 

\usepackage[ruled,vlined]{algorithm2e}
\usepackage{color, soul}
\usepackage{microtype}      
\usepackage{xcolor}         
\usepackage{geometry}
\usepackage{times}
\usepackage{ragged2e}
\usepackage{amsmath}
\usepackage{bm}
\usepackage{latexsym}
\usepackage{arydshln}
\usepackage{graphicx}
\usepackage{float}
\usepackage{longtable}
\usepackage{booktabs} 
\usepackage{multirow}
\usepackage{amssymb}
\usepackage{longtable}
\usepackage{tabu}
\usepackage{bbding}
\usepackage{awesomebox}
\usepackage{bbding}
\usepackage[most,breakable]{tcolorbox}
\usepackage{etoolbox}
\usepackage{fancyhdr}
\usepackage{lipsum}
\usepackage{CJKutf8}
\usepackage{pdfpages}
\usepackage{multicol}
\usepackage{pgffor} 
\usepackage{fontawesome5}
\usepackage{nicematrix} 
\usepackage{paralist}
\usepackage[edges]{forest}
\usepackage{siunitx}
\usepackage{tikz-dependency}
\usepackage{tikz}
\usepackage{bbm}
\usepackage{amssymb}
\usepackage[english]{babel}  
\usepackage{hyphenat}
\usepackage{siunitx}

\newcommand{\Rmnum}[1]{\expandafter\@slowromancap\romannumeral #1@}

\usepackage{caption} 
\usepackage{chngcntr}

\usepackage[colorlinks, linkcolor=RoyalBlue, anchorcolor=BrickRed, citecolor=RoyalBlue, urlcolor=RoyalBlue]{hyperref}

\usepackage{cleveref}
\crefname{section}{§}{§§}
\Crefname{section}{§}{§§}

\newcommand\refsec[1]{Section~\hyperref[sec:#1]{\ref{sec:#1}}}
\newcommand\refsecs[2]{\hyperref[sec:#1]{§\ref{sec:#1}:~\textsc{#1}}, \hyperref[sec:#2]{§\ref{sec:#2}:~\textsc{#2}}}

\usepackage[tikz]{bclogo}
\definecolor{msftBlue}{RGB}{0,164,239}
\definecolor{msftGreen}{RGB}{127,186,0}
\definecolor{msftYello}{RGB}{255,185,0}
\definecolor{mypurple}{RGB}{138,43,226} 
\definecolor{msftBlack}{RGB}{0,0,0}

\newtcolorbox{myboxnote}[1][]{
  breakable,
  title=#1,
  colback=cyan!0,
  colbacktitle=cyan!0,
  coltitle=black,
  fonttitle=\bfseries,
  bottomrule=0pt,
  toprule=0pt,
  leftrule=1.5pt,
  rightrule=1.5pt,
  titlerule=0pt,
  arc=0pt,
  outer arc=0pt,
  colframe=lightgray,
}
\usepackage{mdframed}

\definecolor{academicblue}{RGB}{54, 95, 145}

\tcbset{
  iclrtakeawaybox/.style={
    width=\linewidth,
    colback=gray!5,                 
    colframe=gray!60,              
    colbacktitle=gray!30,            
    coltitle=black,                
    fonttitle=\bfseries,     
    boxrule=0.5pt,                 
    arc=2pt,                       
    sharp corners=south,           
    attach boxed title to top left={yshift=-0.1in,xshift=0.15in},
    boxed title style={boxrule=0pt, colframe=white},
    enhanced,
    drop shadow southeast           
  }
}
\newtcolorbox{TakeawayBox}[2][]{iclrtakeawaybox,title=#2,#1}

\newenvironment{itemsize*}%
 {\leftmargini=20pt\begin{itemize}%
  \setlength{\itemsep}{3pt}%
  \setlength{\parskip}{0pt}%
  }%
 {\end{itemize}}
\newenvironment{enumerate*}%
 {\begin{enumerate}%
  \setlength{\itemsep}{0pt}%
  \setlength{\parskip}{0pt}}%
 {\end{enumerate}}

\usepackage{fancyhdr} 
\usepackage{blindtext} 
\usepackage{ulem}

\usepackage{xparse}
\usepackage{colortbl}

\usepackage{svg}

\title{
\vspace{-2em}
\fontsize{16}{19}\selectfont Vision Enhancing LLMs: Empowering Multimodal Knowledge Storage and Sharing in LLMs}
\author{
Yunxin Li\thanks{Details of authors are shown in Sec. \ref{sec:contributors}. $\ddagger$ indicates the corresponding author}~,  Zhenyu Liu, Baotian Hu$^\ddagger$, Wei Wang, Yuxin Ding, Xiaochun Cao, and Min Zhang \\
Research Institute of Computing and Intelligence\\
{Harbin Institute of Technology, Shenzhen}\\
}

\begin{document}

\maketitle
\vspace{-1em}

\begin{abstract}
 Recent advancements in multimodal large language models (MLLMs) have achieved significant multimodal generation capabilities, akin to GPT-4. These models predominantly map visual information into language representation space, leveraging the vast knowledge and powerful text generation abilities of LLMs to produce multimodal instruction-following responses. We could term this method as \textit{LLMs for Vision} because of its employing LLMs for visual understanding and reasoning, yet observe that these MLLMs neglect the potential of harnessing visual knowledge to enhance the overall capabilities of LLMs, which could be regarded as \textit{Vision Enhancing LLMs}. In this paper, we propose an approach called \textbf{MKS2}, aimed at enhancing LLMs through empowering \textbf{M}ultimodal \textbf{K}nowledge \textbf{S}torage and \textbf{S}haring in LLMs. Specifically, we introduce Modular Visual Memory (MVM), a component integrated into the internal blocks of LLMs, designed to store open-world visual information efficiently. Additionally, we present a soft Mixture of Multimodal Experts (MoMEs) architecture in LLMs to invoke multimodal knowledge collaboration during text generation. Our comprehensive experiments demonstrate that MKS2 substantially augments the reasoning capabilities of LLMs in contexts necessitating physical or commonsense knowledge. It also delivers competitive results on image-text understanding multimodal benchmarks. The codes will be available at: \url{https://github.com/HITsz-TMG/MKS2-Multimodal-Knowledge-Storage-and-Sharing}
\end{abstract}



\section{Introduction}
\label{sec:introduction}

Recent advances~\citep{yin2023survey,driess2023palm,li2023lmeye,ye2023mplug} on multimodal large language models (MLLMs) have opened the eyes of text-only large language models (LLM, ``blind'' to visual information), allowing them to understand and process multimodal information, thereby promoting the further development of LLMs-centered Artificial General Intelligence (AGI). In this line of works, such as MiniGPT-4~\citep{zhu2023minigpt}, LLaVA~\citep{liu2023visual}, and BLIP-2~\citep{li2023blip2}, information outside language modality is usually aligned into the language space. Then, the rich knowledge stored in LLM and its powerful text generation capability are used to understand various multimodal information and generate the response corresponding to human instructions.
They took a significant step towards constructing a multimodal large visual-language model similar to GPT-4~\citep{gpt4}, contributing a lot of multimodal instruction-following data~\citep{zhang2023llavar, liu2023visual, liu2023mmbench} and efficient multimodal fine-tuning technical~\citep{ye2023mplug, zhu2023minigpt}. These approaches, concentrating on multimodal information understanding, could be regarded as \textit{``LLMs for vision''}, primarily because they mainly utilize LLMs for processing visual understanding problems. 

\begin{figure}[t]
    \centering
    \includegraphics[width=0.85\columnwidth]{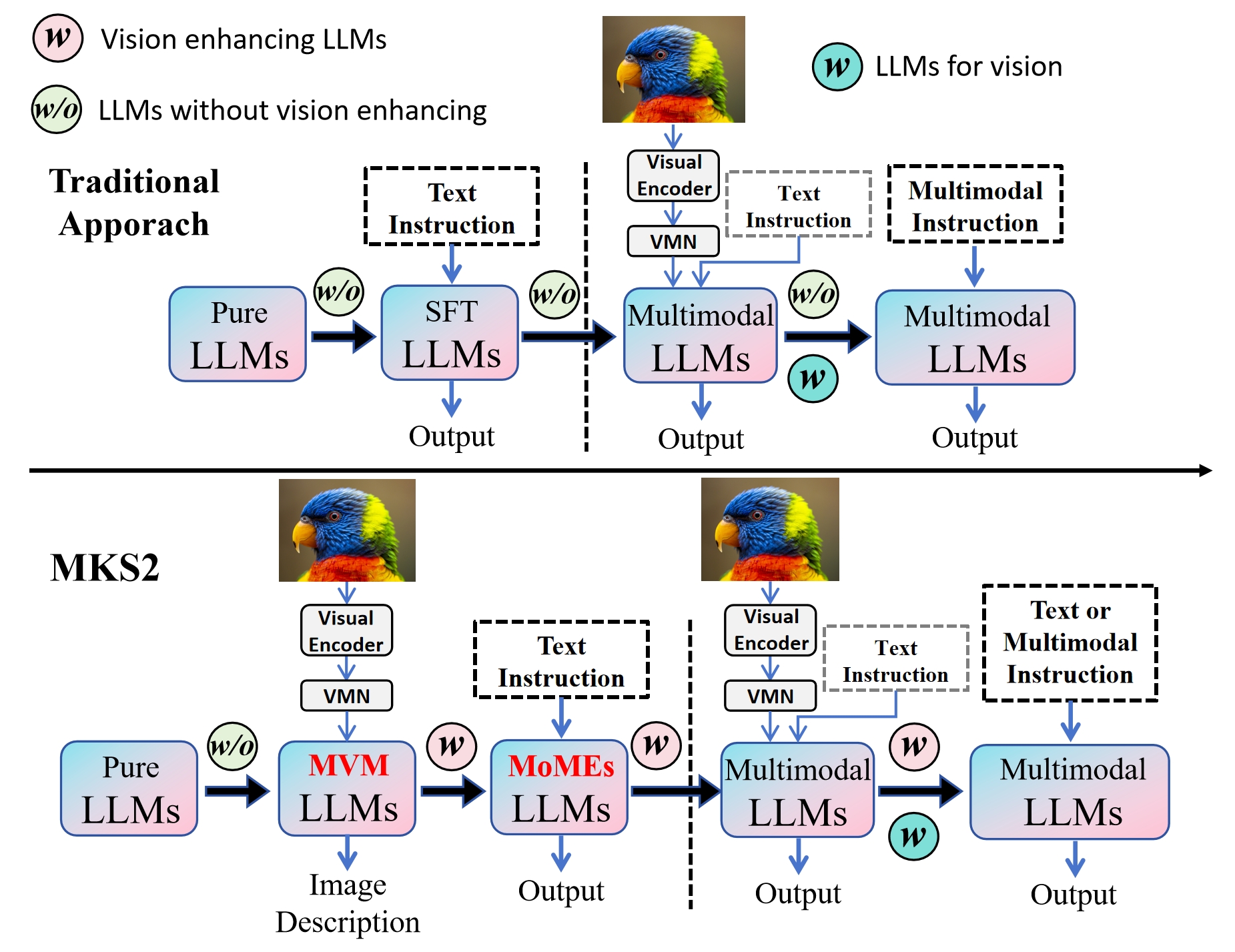}
    \caption{Comparisons between the proposed MKS2 and previous supervised fine-tuned (SFT) and multimodal LLMs. MKS2 focuses on improving LLMs with visual knowledge. VMN refers to the visual mapping network, transferring image encoding to the language space. MVM and MoMEs represent the proposed modular visual memory and the architecture of a soft mixture of multimodal experts in LLMs, respectively.}
    \label{fig:intro_case}
\end{figure}

However, whether current MLLMs or pretrained and Supervised Fine-Tuned (SFT) LLMs overlook enhancing the ability of LLMs to tap into visual knowledge. Most existing multimodal models are mainly designed for vision-language understanding, focusing on interpreting or reasoning about a given image in response to a text query. They cannot acquire new, generalizable knowledge from visual data to answer future text-only questions. Similarly, text-only LLMs only learn, store, and apply knowledge from textual data. Yet, some visual knowledge cannot be fully captured by language alone, and learning from visual experience may enhance both language and object understanding in LLMs. 
Ideally, just as the human brain retains and utilizes visual information, MLLMs or LLMs should be equipped to store external visual information. In situations that require visual common sense, even in the absence of direct visual input, LLMs should be able to access this stored visual-language knowledge for combined reasoning. This goes beyond merely processing multimodal input, as ``LLMs for Vision'' depicted in the top part of Figure~\ref{fig:intro_case}. Hence, we first present a term \textit{``Vision Enhancing LLMs"} to describe the desired capability for LLMs. Through this enhancement, large models would store and effectively draw upon multimodal knowledge, and their knowledge base and reasoning capabilities would be enhanced.

To this end, we present \textbf{MKS2}, an innovative approach designed for empowering \textbf{M}ultimodal \textbf{K}nowledge \textbf{S}torage and \textbf{S}haring within LLM, consisting of two core stages: Visual Information Storage and Multimodal Knowledge Collaboration.
In the first stage, we introduce Modular Visual Memory (MVM) in internal transformer blocks of LLMs to store visual information. Specifically, inspired by previous works~\citep{kazemnejad2023measuring, wang-etal-2022-finding-skill} focused on measuring parametric knowledge of pretrained language models and observing the knowledge storage role of feed-forward neural networks (FNN), we incorporate two layers of FNN into each LLM block to build a lightweight visual memory. Subsequently, we employ a collection of image-text pairs to exclusively train and update MVM using two learning approaches: image-to-text generation and text-to-image retrieval or generation. The other parameters of pretrained LLMs will be frozen during training. In both ways, soft image and token embeddings pass through the visual memory following attention calculations. These learning strategies mainly empower LLMs to comprehend, translate, and store visual information via visual memory in LLMs and with a vision-centric learning approach.

For multimodal knowledge collaboration, we introduce a soft Mixture of Multimodal Experts (MoMEs) architecture to recall stored visual and language knowledge. This framework leverages specialized experts, including the pretrained Modular Visual Memory (Visual Expert) and the originally trained MLPs (Textual Expert) in LLMs during the generation process. To efficiently achieve this, we freeze all parameters of LLMs, apply Low-Rank Adaption (LoRA~\citep{hu2021lora}) to each expert module, and facilitate information integration across LLM blocks through a token-level soft mixing approach. By doing so, the overall model becomes adept at accommodating both multimodal and text-modality information, enabling seamless collaboration across various input forms. During training, we collect a diverse set of instruction data, containing text-only instructions and image-text multimodal instruction-following data, to ensure the effectiveness of MoMEs in handling multimodal as well as text-only tasks. 

\begin{wrapfigure}{r}{0.5\textwidth}  
    \centering
    \includegraphics[width=\linewidth]{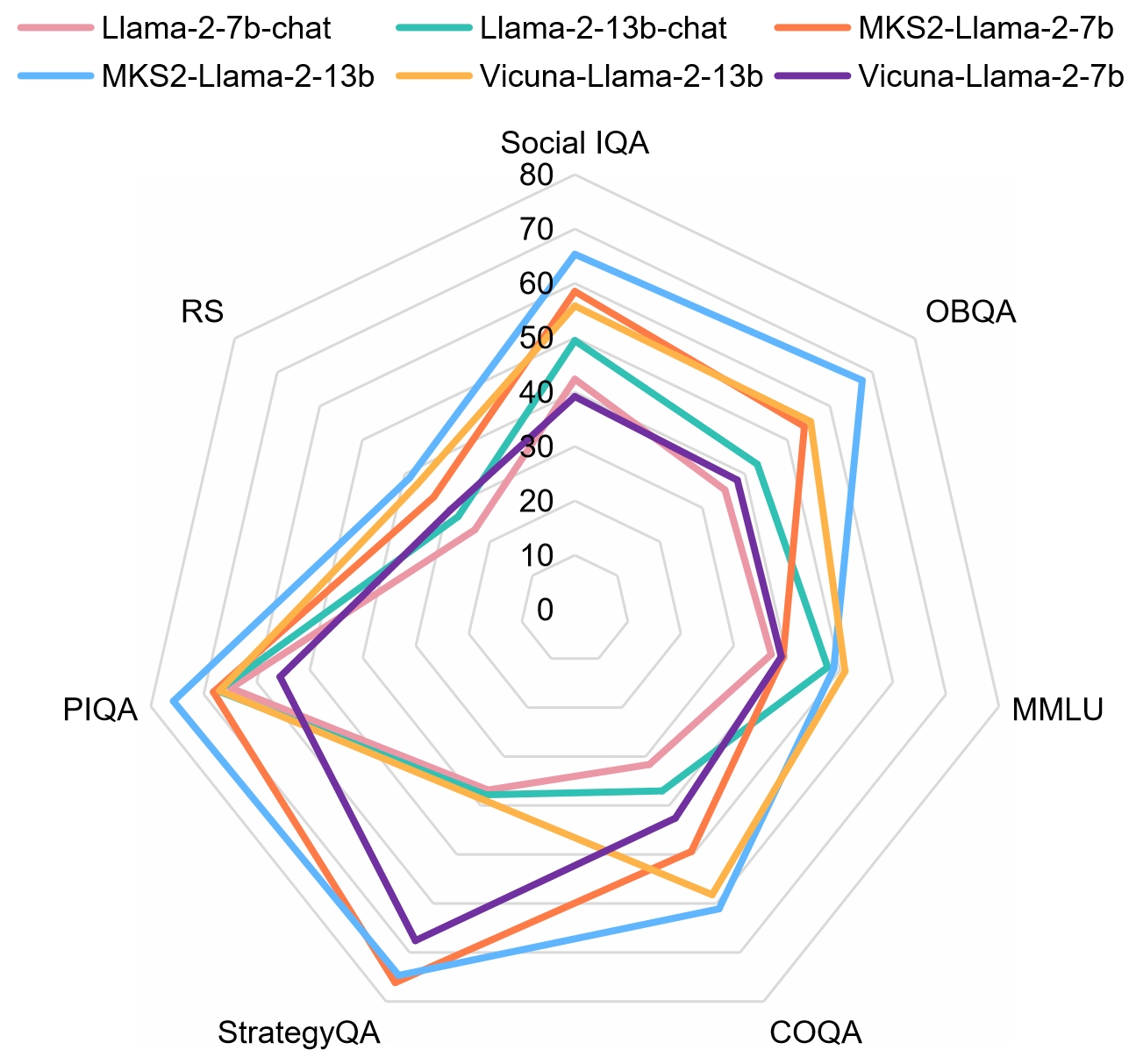}
    \caption{MKS2-Llama-2-13b achieves SOTA zero-shot performance on seven natural language reasoning tasks. It indicates that achieving multimodal knowledge storage and sharing is effective for improving LLMs.}
    \label{fig:leida}
\end{wrapfigure}

A fundamental distinction sets MKS2 apart from mainstream MLLMs, which we characterize as a shift from ``LLMs for Vision" to a new paradigm we term ``Vision-Enhanced LLMs." This distinction is not merely architectural but is rooted in a divergent primary objective. Traditional MLLMs (e.g., BLIP-2, LLaVA) aim to equip LLMs with visual understanding, typically evaluated on tasks like VQA. In this framework, the image is the subject of the query, and the LLM acts as a visual interpreter. While this approach has proven highly effective for multimodal benchmarks, it often leads to a well-documented trade-off: a degradation of the model's pure language capabilities. This occurs because the intensive process of vision-language instruction tuning can distort the model's original linguistic representations, as optimization is squarely focused on cross-modal alignment. In contrast, MKS2 is not merely to process images, but to use visual information as a source to enhance and enrich the LLM's own internal knowledge base and linguistic capabilities. Here, the image is not the end goal but a source of knowledge from which the model learns. The objective is to absorb information from the visual world and integrate it into the LLM's parametric memory in a language-compatible format. This allows the model to solve knowledge-intensive text-based questions that it could not answer before seeing the relevant visual data, much like a human ``learning with their eyes open". While ``LLMs for Vision" teach a model to see and describe, our ``Vision-Enhanced LLM" paradigm teaches it to see and know.

To validate the effectiveness of our approach, we evaluate MKS2 on seven natural language processing (NLP) benchmarks and six image-text understanding datasets. Extensive experiment results indicate that MKS2 achieves superior performances on NLP tasks requiring physical or visual world knowledge, e.g., MKS2-Llama-2 significantly exceeds Llama-2-chat. It also achieves competitive performances on image-text understanding scenarios compared to MLLMs using the same supervised fine-tuning data. 

Our main contributions can be summarized as follows:

\begin{itemize}
    \item We introduce MKS2, a vision-enhanced learning framework for LLMs, designed for effective storage and sharing of multimodal knowledge. This framework efficiently handles both multimodal and text-only inputs.

    \item MKS2 demonstrates superior outcomes in knowledge-intensive tasks over traditional SFT LLMs and LLMs employing Reinforcement Learning from Human Feedback (RLHF), as the results shown in Figure~\ref{fig:leida}.
    
    \item Ablation studies validate the efficacy of a mixture of multimodal experts that incorporate visual and language knowledge experts. This architecture distinctly improves the performance of LLMs with a few training parameters ($< 0.25$\% of LLM parameters).

    \item Our analysis also indicates that the multimodal instruction-following data could enhance LLMs' performance in commonsense reasoning tasks and text-only question-answering data can improve the open-ended visual question-answering tasks.  

\end{itemize}

\section{Related Work}
\label{sec:related_work}

\subsection{Large Language Model}

A large language model (LLM) serves as a sophisticated computational framework specifically crafted to generate human-like language and execute a wide array of natural language processing tasks. These models~\citep{lmopenai,NEURIPS2022_b1efde53}, such as GPT-4~\citep{gpt4} and LLaMA~\citep{gao2023llama}, acquire their extensive capabilities by meticulously analyzing vast text datasets through advanced training methods, including self-supervised and semi-supervised learning techniques~\citep{Zhai_2019_ICCV}. The pinnacle of LLM technology is represented by models employing a decoder-only transformer architecture~\citep{vaswani2017attention}, which optimizes both text processing and generation efficiency. Additionally, modern LLMs can be tailored for specific use cases or further enhanced through prompt engineering~\citep{promptliu}, allowing users to achieve desired outputs more effectively. The training target~\citep{gloeckle2024better} for large language models is predicting the next token that follows a given sequence of tokens. By iteratively making predictions and adjusting its internal parameters based on the differences between its predicted tokens and the actual next tokens in the training data, the model learns to capture complex patterns and structures inherent in language. While the primary training corpus~\citep{fang2024large,NEURIPS2023_9d89448b} for these models predominantly comprises textual data sourced from the Internet, much of this text has already been utilized in the pertaining stage. However, it is important to recognize that images, videos, and other visual mediums~\citep{10368165,10375886,10418849} possess a wealth of common-sense knowledge that remains largely untapped. This visual information presents an opportunity to significantly augment the performance and capabilities of large language models, especially for expanding the knowledge storage of LLM. In this paper, we focus on exploring how to leverage visual knowledge derived from images to enhance the overall performance of LLMs, ultimately aiming to bridge the gap between textual and visual understanding in artificial intelligence.

\subsection{Visual knowledge enhanced methods}

The integration of explicit visual information has been a significant area of research aimed at enhancing the imaginative representation capabilities of language models, ultimately fostering their diverse generative potential. A notable contribution in this field is by \citep{jin-etal-2022-leveraging,10704586}, who investigates how visual knowledge can be harnessed within natural language processing tasks. They develop various cross-model enhancement techniques designed to bolster the representation abilities of pretrained language models, paving the way for richer and more contextually aware text generation. Several other studies \citep{shi-etal-2019-visually, lu2022imagination, li2023multi, 10480354} have explored the approach of retrieving relevant images~\citep{9858662} from an image corpus based on textual input. The retrieved images will be fed into the language model as supplementary information. This method effectively utilizes visual knowledge to enhance performance on downstream tasks, including text completion \citep{zellers-etal-2019-hellaswag}, story generation \citep{fan-etal-2018-hierarchical}, and concept-to-text transformations \citep{barzilay2005collective}. Recently, researchers \citep{long-etal-2021-generative, yang2021open, zhu2022visualize} have introduced innovative techniques that utilize advanced text-to-image technologies to create imaginative representations of language. These representations are then integrated into language models through prefix-tuning~\citep{li-liang-2021-prefix} methodologies. Historically, approaches for enhancing language models with visual information have primarily focused on smaller models, with many methods relying on retrieval-augmented pipelines to incorporate visual knowledge. In this paper, we aim to investigate the concept of visual information storage within large language models and demonstrate how these models can be enhanced with visual knowledge without the need for explicit image inputs.

\subsection{LLMs for vision}

Recent works~\citep{zhang2023llavar, zhu2023minigpt, li2023blip2} towards multimodal LLMs focus on utilizing the extensive knowledge and language generation capabilities of LLMs to solve multimodal tasks, especially for visual understanding and reasoning. The construction of these multimodal LLMs mainly contains the following steps. Firstly, these works usually map the visual information obtained by a pretrained visual encoder into the representation space of LLMs, through a learnable linear projection layer~\citep{merullo2022linearly}, MLP, or Q-Former~\citep{li2023blip2}. This stage is usually called visual-language feature alignment and only a few data may be needed to do a good job~\citep{liu2023visual}. Secondly, the initial MLLMs will be tuned via multimodal instruction-following data~\citep{ye2023mplug,li2023lmeye,bai2023qwen}. At this stage, LLMs and the projection layer are often tuned together only with multimodal instruction data. The commonly used LLM is the SFT-LLM tuned with the lightweight LoRA~\citep{hu2021lora}. 

In parallel, advances in instruction-tuning text-only LLMs have demonstrated remarkable performance in NLP tasks and human-machine interactions, as seen in models like Flan-T5, Bloomz~\citep{muennighoff2022crosslingual}, and ChatGPT~\citep{ouyang2022training, muennighoff2022crosslingual, chung2022scaling}. 
Recently, some researchers have explored using multimodal instruction data~\citep{liu2023visual,zhang2023llama,li_unimoe} to fine-tune pre-trained LLMs to improve their multimodal human-machine interaction capability. \citep{liu2023visual} employs the GPT-4 to produce the multimodal instruction data and fine-tune the language model LLaMA~\citep{touvron2023llama} on the synthetic multimodal instruction-following dataset. \citep{zhu2023minigpt}, \citep{li2023monkey}, \citep{cha2023honeybee}, and \citep{wang2023all} also introduce a well-aligned multimodal instruction-following dataset to fine-tune a robust instruction-tuned language model (Vicuna). 
Recent advanced MLLMs usually achieve superior performance on open-domain multimodal question-answering tasks by integrating more powerful language models and larger, higher-quality instruction datasets, as demonstrated by models like LLaMA-VID~\citep{li2023llamavid}, Qwen-VL~\citep{bai2023qwen}, MiniGPT-5~\citep{zheng2023minigpt5}, InternLM-XComposer2~\citep{dong2024internlm} and MobileVLM V2~\citep{chu2024mobilevlm}. 
Moreover, some researchers are focusing on building MLLMs capable of processing high-resolution images while ensuring high inference efficiency and safety, as explored in studies such as Satety Fine-tuning~\citep{zong2024safety}, MoE-LLaVA~\citep{lin2024moe}, and LLaVA-NeXT~\citep{liu2024llavanext}.
In conclusion, these works emphasize leveraging the textual knowledge of large language models to enhance visual understanding and reasoning, contributing to the development of robust MLLMs.

\section{Preliminaries}


\subsection{Supervised Fine-tuning}

A pure pretrained large language model is often fine-tuned on high-quality labeled datasets using token-level supervision to produce a Supervised Fine-Tuned model, dubbed SFT-LLM. Common methods are using GPT-4 automatically constructed instruction data~\citep{selfinstruct} and manually annotated high-quality data from downstream tasks~\citep{chung2022scaling} to fine-tune pure LLMs. Recent works present efficient instruction-tuning approaches to reduce training costs, e.g., LoRA~\citep{hu2021lora}, QLoRA~\citep{dettmers2023qlora}, and LoftQ~\citep{li2023loftq}. These SFT-LLMs are capable of generating human-like responses for various text-only instructions, having a profound impact on all walks of life.  

\begin{figure*}[t]
    \centering
    \includegraphics[width=0.90\textwidth, height=0.45\textwidth]{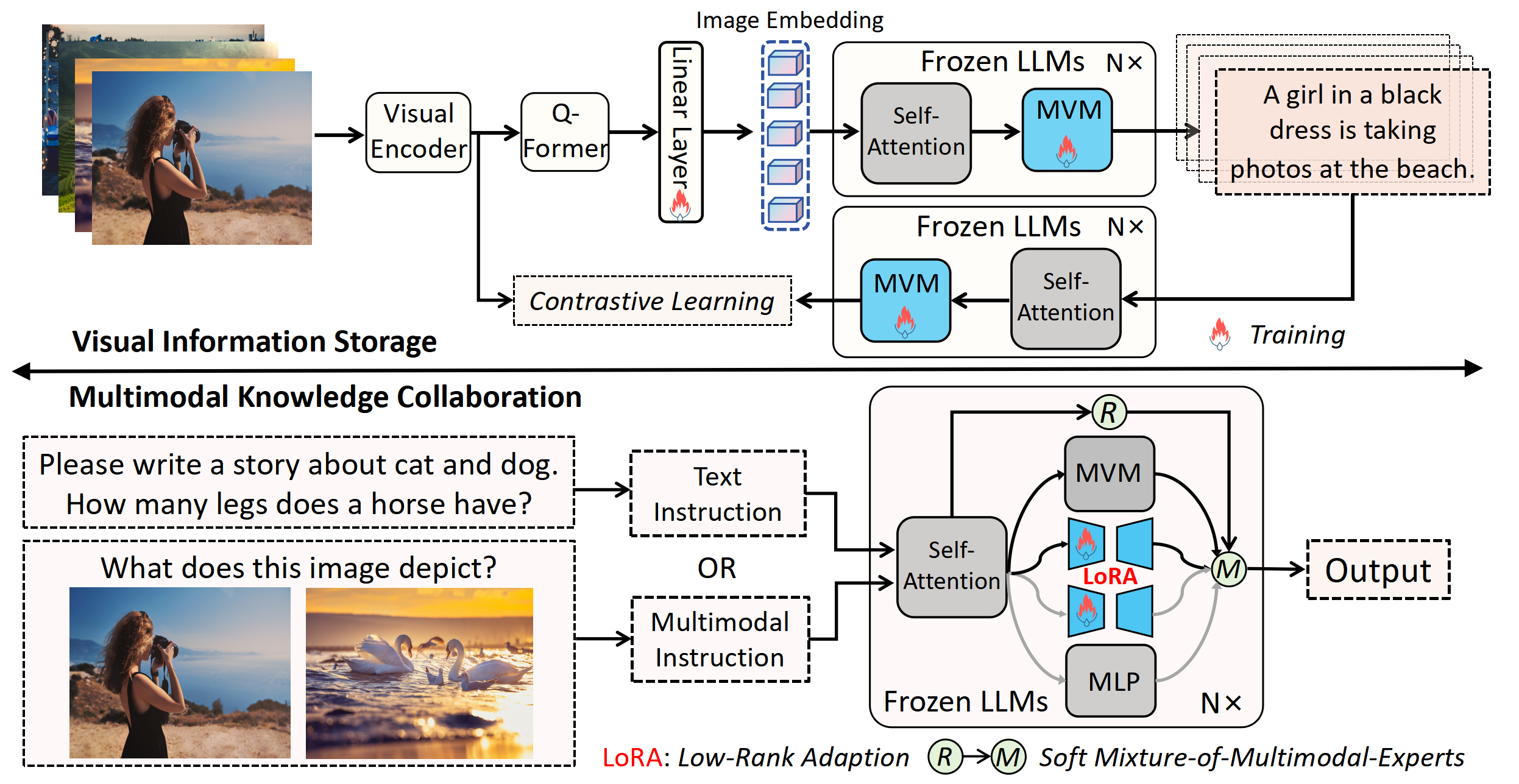}
    \caption{The overall workflow of MKS2. It realizes visual information storage and multimodal knowledge collaboration in LLMs. In the first stage, we introduce the modular visual memory (MVM) and train it through language-centric learning strategies on large-scale image-text pairs. We also present a soft mixture-of-multimodal experts (MoMEs) architecture to accomplish multimodal knowledge collaboration during text generation. }
    \label{fig:model}
\end{figure*}

\subsection{Multimodal Instruction-Following Tuning}

Compared to traditional visual-language models such as Oscar~\citep{li2020oscar}, Flamingo~\citep{alayrac2022flamingo}, OFA~\citep{wang2022unifying}, etc, the multimodal instruction-following tuning approach explored extending the text-only instruction tuning in LLMs to multi-modality. These MLLMs applying LLMs as the multimodal information processor achieve impressive zero-shot performances on unseen tasks. Generally, as the traditional approach depicted in Figure~\ref{fig:intro_case}, a frozen visual encoder (e.g., the visual encoder of CLIP) is used to obtain the sequence representation of an image and a visual mapping network (VMN, a linear projection layer or Q-former from BLIP-2) projects the image encoding into soft image embeddings into the language space of LLMs. Then, we can utilize an efficient fine-tuning technique to allow LLMs to process multimodal information, thereby turning LLMs into MLLMs. 

Formally, a multimodal image-text instruction sample could be expressed in the following triplet form, i.e., $(I,T,R)$, where $I, T, R$ represent the input image, text description (about human demands or image-related premises), and ground-truth response, respectively. During training, the constructed MLLMs are forced to predict the next token of response via the autoregressive objective, which could be presented as:
\begin{equation}
    \mathcal{L}(\theta)=-\sum_{i=1}^N \log P\left(R_i \mid I, T, {R}_{<i} ; \theta\right),
\end{equation}
where $N$ is the length of response and $\theta$ refers to the training parameters in the whole framework. 

In conclusion, we find that these two approaches overlook introducing visual knowledge to improve the overall capabilities of LLMs for processing text-only tasks.

\section{Methodology}

In the following subsections, we will present the two stages of MKS2 in detail: Visual Information Storage and Multimodal Knowledge Collaboration.

\subsection{Visual Information Storage}

To realize visual information storage in LLMs, we propose injecting Modular Visual Memory (MVM) into internal blocks of LLMs and forcing MVM to memorize open-world visual information via language-centered learning strategies. 

\textbf{Modular Visual Memory (MVM)}. This module is a two layer of feed-forward neural networks (FFN) and is injected into each transformer block of LLMs. As the top part shown in Figure~\ref{fig:model}, the input image $I$ is first projected into soft image embedding $\mathbf{h}_I$ via the pretrained visual encoder (CLIP~\citep{hafner2021clip}), Q-former from BLIP-2 or a simple MLP like the LLaVA~\citep{liu2024llavanext}, and a learnable linear layer. Take the first block as an example; the calculation process can be presented as follows:
\begin{equation}
    \begin{aligned}
& \mathbf{h}_s^T=\text{Self-Attention}\left(\mathbf{h}_I\right), \\
& \mathbf{h}_F^T=\mathbf{h}_s^T + \mathrm{MVM}\left(\text{layernorm}(\mathbf{h}_s^T)\right), \\
\end{aligned}
\label{eq1}
\end{equation}
where $\text{Self-Attention}$ is the original attention calculation in LLMs. We just inserted MVM inside the original LLMs and did not change other structures. All hidden states pass the MVM after gaining the output $\mathbf{h}_s^T$ of $\text{Self-Attention}$, and we also set the overall size of visual memory by controlling the hidden state dimensions of FFN.

\textbf{Language-Centered Learning Strategies}. As we consider LLMs as analogues to the human brain, we have embarked on a groundbreaking endeavour to create visual storage memory in LLMs. Our ultimate goal is to empower LLMs with the capability to comprehend a given image and conjure related visual scenarios based on textual input, akin to human cognition. To this end, we adopt two learning objects to train MVM with a large amount of image-text pairs. As the top part is shown in Figure~\ref{fig:model}, we allow LLMs to generate the language description of an image, which resembles understanding and translating an image like a brain. Additionally, given a sentence with some visual objects, LLM should attach to the sentence-related image, which resembles the imagination ability of a human. Suppose that the short description (caption) of an input image $I$ is $D$, the description generation loss is
\begin{equation}
    \mathcal{L}_c=-\frac{1}{N} \sum_{i=1}^N l_c\left(IMG_i, D_i\right),
\end{equation}
where $N$ is the number of image-text pairs in a batch and $l_c$ refers to the cross-entropy loss. At this stage, the visual token of the image and the text token of the caption are all passed through the newly added Visual MLP in each layer of LLM.

While retrieving the related image from sentences, we use the output hidden state $h_e$ of the end token $\left</s\right>$ of the input caption to match the image embedding. Concretely, we employ a learnable linear layer to project it into the same dimension with image global encoding obtained by the visual encoder. Then we calculate the cosine similarity between them and minimize the InfoNCE loss for text-to-image (t2i) retrieval over a batch of $N$ samples. The negatives are other irrelevant images in a batch. Hence, the total language-centered learning loss is 
\begin{equation}
\begin{aligned}
     &\mathcal{L}_{Stage 1} = \mathcal{L}_c +  \mathcal{L}_{t2i},\\
      &\mathcal{L}_{t2i} =-\frac{1}{N} \sum_{i=1}^N\left(\log \frac{\exp \left(\operatorname{sim}\left(D_i, \textit{IMG}_i\right) / \tau\right)}{\sum_{j=1}^N \exp \left(\operatorname{sim}\left(D_i, \textit{IMG}_j\right) / \tau\right)}\right),\\
\end{aligned}
\end{equation}
where $\tau$ is a learnable temperature parameter. During training, we freeze all pretrained parameters of LLMs and only update MVM.
In addition to the way of retrieving images to achieve visual information association, using image generation technology for joint training is also an alternative approach.

\begin{wrapfigure}{r}{0.5\textwidth}
    \centering
    \includegraphics[width=0.5\textwidth]{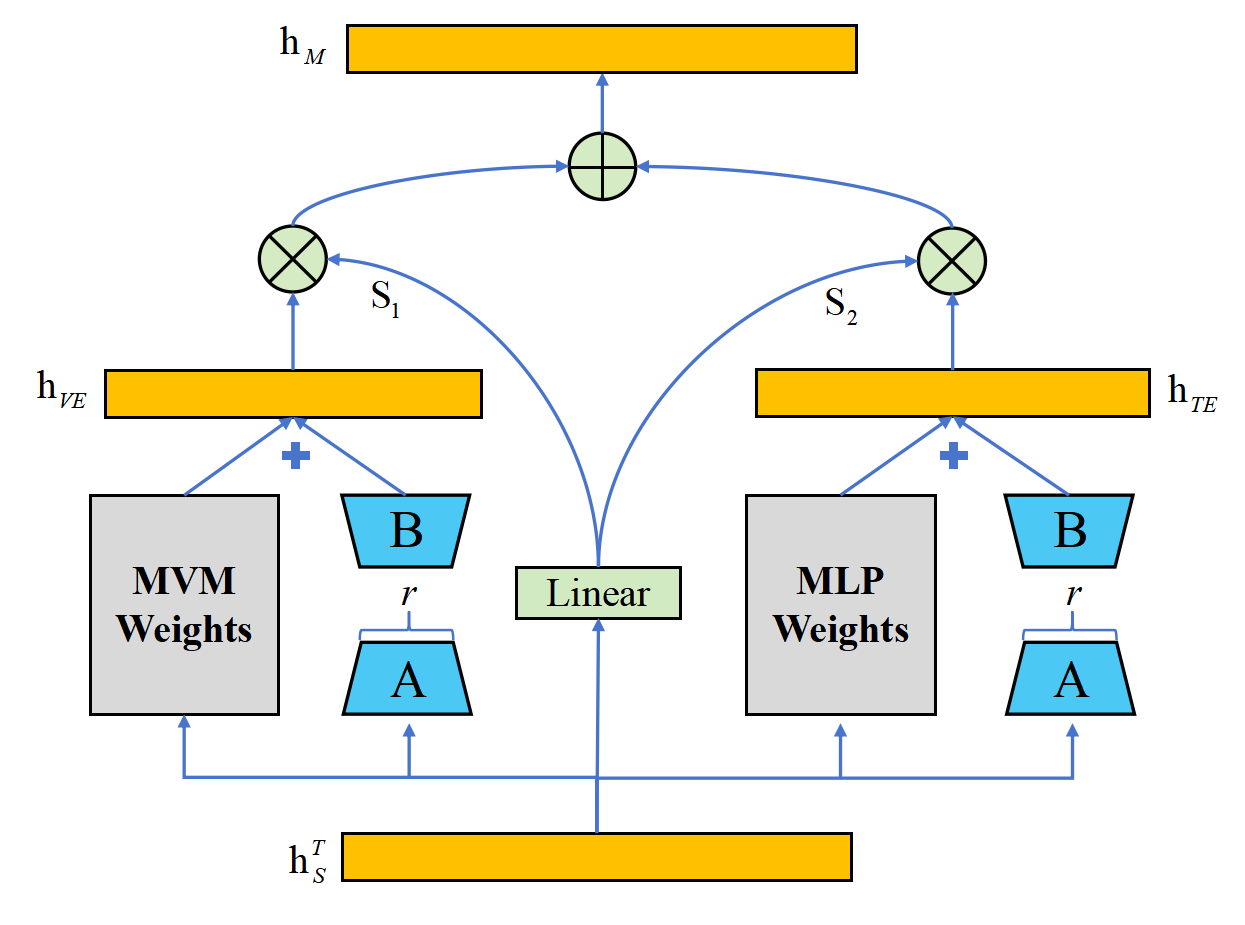}
    \caption{The detailed calculation process of the proposed soft mixture-of-multimodal experts (MoMEs) architecture. It aims to realize multimodal knowledge collaboration during text generation.}
    \label{fig:momoes}
\end{wrapfigure}

\subsection{Multimodal Knowledge Collaboration}

After gaining visual information storage inside LLMs, we need to consider how to realize multimodal knowledge collaboration during generation. Regarding pretrained MVM and MLP in LLMs as visual and textual experts, respectively, we propose a soft mixture-of-multimodal experts approach to achieving multimodal knowledge utilization at the token level.

\textbf{Mixture-of-multimodal Experts (MoMEs)}. To speed up the training process, as the bottom part shown in Figure~\ref{fig:model}, we freeze MVM and other parameters of LLMs, applying Low-Rank Adaption~\citep{hu2021lora} (LoRA) to tow-modality experts: MVM and MLP. We denote the tokens of the input for one sequence inputted to MoMEs by $\mathbf{X} \in \mathbb{R}^{m \times d}$, where $m$ is the number of tokens and $d$ is their dimension. As show in Figure \ref{fig:momoes}, the computed process for visual and language knowledge experts could be given in
\begin{equation}
    \begin{aligned}
        & \mathbf{h}_{VE} = \textit{LoRA-MVM}(X), \\
        & \mathbf{h}_{TE} = \textit{LoRA-MLP}(X), \\
        & LoRA(W_0) := W_0 X+\Delta W X=W_0 x+B A X, \\
    \end{aligned}
\label{eqlora}
\end{equation}
where $B, A$ are learnable parameters added for each pretrained weight of the visual and textual experts. \textit{LoRA-MVM}(X) and \textit{LoRA-MLP}(X) represent original knowledge experts equipped with additional LoRA calculation. By adopting this method, the training process becomes more efficient, as it does not require updating the overall parameters of two experts.

Each MoE layer uses expert functions (shown in E.q.~\ref{eqlora}) applied on individual tokens, namely $\left\{f_i: \mathbb{R}^d \rightarrow \mathbb{R}^d\right\}_{1: 2}$. Each expert will process $p$ slots, and each slot has a corresponding $d$-dimensional vector of parameters. As \textit{R}$\rightarrow$\textit{M} shown in Figure~\ref{fig:model}, the token-level combination for expert outputs can be presented as
\begin{equation}
    \begin{aligned}
        & S = Softmax(w_s X + b_s), \\
        & \mathbf{h}_M = S_1 \mathbf{h}_{VE} + S_2 \mathbf{h}_{TE},\\
    \end{aligned}
\end{equation}
where $S\in R^{X\times2}$ refers to the weights of each expert, and the final dimension is normalised with Softmax calculation. The output of the MoMEs block in LLMs is denoted to $\mathbf{h}_M$.

\section{Training and Data Recipes}
\label{sec:training_data}

In the first stage, the size of image-text pairs is about 2.3M (Million) from CC3M~\cite{changpinyo2021conceptual}, COCO Captioning~\cite{chen2015microsoft},  and Flick-30k~\cite{plummer2015flickr30k}. 
To achieve multimodal knowledge collaboration, as shown in Figure~\ref{fig:intro_case}, we use text-only and image-text instruction-following data to train the overall architecture. The modular visual memory and LLMs are frozen during training. We use widely-used instruction data including natural language processing tasks filtered from Flan-T5~\cite{chung2022scaling}, complex instruction-finetuning data from WizardLLMs~\cite{xu2023wizardlm}, and multimodal instruction data LLaVAR~\cite{zhang2023llavar}, which consists of 1.5M textual and 166k image-text question-answering pairs. 

During text-only inference, the model functions as a powerful LLM, with the added capability of recalling knowledge stored in its MoMEs from past multimodal learning. The transition between multimodal and text-only inputs is seamless because the visual processing pathway is an additive, optional module that leaves the core language model intact. In this mode, the MoMEs remain dormant yet accessible, having been integrated into the LLM's architecture and imbued with knowledge from visual experiences. When faced with a question that requires visual knowledge, e.g., ``What is the Eiffel Tower made of ?'', the model can activate the relevant experts within the MoMEs to retrieve the necessary information, analogous to how it activates neurons for purely textual facts. For multimodal inputs, our model functions similarly to traditional MLLMs, utilizing a vision encoder to process the input image.

\section{Experiments}
\label{sec:experiments}

\subsection{Datasets}

\subsubsection{Natural Language Processing Benchmarks} We use seven text-only downstream datasets to comprehensively evaluate MKS2, which consists of physical world knowledge-relevant datasets and basic ability assessment benchmark MMLU~\citep{hendryckstest2021}. We use multiple choice question answering tasks that can benefit from visual knowledge: PIQA~\citep{Bisk2020} that requires physical commonsense reasoning, Commonsense QA (CSQA)~\citep{talmor-etal-2019-commonsenseqa} for evaluating the commonsense reasoning capability of models, OpenBook QA (OBQA)~\citep{OpenBookQA2018} that requires multi-step reasoning, use of additional common and commonsense knowledge, and rich text comprehension, RiddleSense (RS)~\citep{lin2021riddlesense} for complex understanding of figurative language and counterfactual reasoning skills, Social IQA~\citep{sap2019socialiqa} which focuses on physical or taxonomic knowledge for testing social commonsense intelligence, StrategyQA~\citep{geva2021strategyqa} that needs the reasoning steps should be inferred using a strategy.

\subsubsection{Image-Text Understanding Benchmarks}
To evaluate the multimodal capability of our proposed model, we introduce six classical Visual Question Answering (VQA) datasets: VQAv2~\citep{VQA}, OK-VQA~\citep{marino2019ok}, ST-VQA~\citep{biten2019scenevqa}, OCR-VQA~\citep{OCRVQA}, TextVQA~\citep{STVQA}, and DocVQA~\citep{mathew2021docvqa}. VQAv2 is a classic open-world VQA dataset, containing more than 1 million samples. Scene Text Visual Question Answering (STVQA) consists of 31,000+ questions across 23,000+ images collected from various public datasets. The OCRVQA dataset includes more than 1 million question-answer pairs that cover over 207,000 book cover images. The TextVQA dataset consists of over 45,000 questions related to text on more than 28,000 images selected from specific categories of the OpenImages dataset. DocVQA is a comprehensive dataset comprising 12,767 document images with diverse types and content, accompanied by over 50,000 questions and answers.
For datasets containing
far more than 5000 image-question pairs, we selected the first 5000 pairs for the test, similar to \citep{liu2023hidden}.

\subsection{Comparing Models}
The comparing models mainly comprise three types of open-source Llama-2 variants: \textit{SFT Llama-2, RLHF-tuned Llama-2, and recently proposed MLLMs}. To verify the vision-enhancing method MKS2-Llama, we present text-only instruction-tuned variant Llama-2-7b-INST-LoRA and Vicuna-Llama-2~\citep{refinedweb}, where we adopt the textual instruction data identical to our approach and set $r=16$ for LoRA to train Llama2-7b-INST. Hence, the training parameters of Llama-2-7b-INST-LoRA are similar to the proposed MKS2-Llama-2-7B, about 14M.
RLHF-tuned models are language models that are trained using a combination of human feedback and reinforcement learning techniques, achieving better performance in understanding human instructions and generating high-quality responses. Additionally, to evaluate the multimodal information processing capability of MKS2-Llama, we also introduce recent MLLMs as baselines. Flamingo~\citep{alayrac2022flamingo} and OFA~\citep{wang2022unifying} are traditionally pretrained visual-language models, which have seen an amount of image-text pairs. BLIP-2~\citep{li2023blip2} is a widely-used visual and language model, achieving remarkable zero-shot performance on downstream image-text understanding tasks. MiniGPT-4~\citep{zhu2023minigpt}, KOSMOS-2 \citep{peng2023kosmos}, ImageBind-LLM \citep{han2023imagebind}, FROMAGe~\citep{koh2023grounding}, mPLUG-Owl~\citep{ye2023mplug}, LLaVR~\citep{zhang2023llavar} and InstructBLIP~\citep{dai2024instructblip} are multimodal instruction-tuned MLLMs, trained with enormous image-text instruction-following data. In addition, for the knowledge-based VQA tasks, we introduced two task-specific model: ViperGPT \citep{suris2023vipergpt} with code-based visual reasoning and REVEAL \citep{hu2023reveal} with multimodal knowledge retrieval generation.


\subsection{Implementation Details}
We take the pretrained Llama-2 version~\citep{touvron2023llama} as the backbone of MKS2 and run all models with Adam Optimizer~\citep{kingma2014adam} on 4 A100-80G GPUs with a Python environment. All models are trained and tested with the Bfloat16 floating-point format. The dimension of the middle layer of the inserted visual memory module is 1/4 of the hidden state size of LLMs. For Llama-2-7b, the total parameters of MVM are about 410 million. During visual information storage, we take the frozen visual encoder and Q-former from BLIP-2-FlanT5-xxl to obtain the image encodings, so the length of soft image embedding is 32. Additionally, we set the initial learning rate to 1e-4 and trained the model for about 2 epochs with warm-up steps equaling 5000. The batch size is set to 32 with four-step gradient accumulation for a single GPU device. While performing instruction-following learning, we set the batch size, $r$ in LoRA to 3 and 8, respectively, and the max length of the input is set to 1024. To tag the position of image embedding, we introduce two learnable tokens $\left<img\text{-}start\right>$ and $\left<img\text{-}end\right>$. 
Similar to Llama-2-chat, we add [INST] and [/INST] at the start and end of text instruction, as like ``\textit{[INST] Please write a short story about cat and dog [/INST]}''.
During generation, we set beam sizes to 1 and 4 for text-only and VQA tasks, respectively.

\subsection{Overall Performance}
\textbf{Performance of vision-enhancing LLMs}. We present zero-shot model performances in Table~\ref{text_results}, aiming to evaluate the instruction-understanding and open-world problem-solving abilities of LLMs. We observe that the proposed method MKS2-Llama-2-7B/13B achieves the best performance on almost all evaluation datasets, especially on substantially suppressing Llama-2-7b/13b-chat. Compared to powerful Llama-2-7b-INST-LoRA of the same magnitude, MKS2-Llama-2-7b could gain by about \textbf{8}\% on CommensenseQA, \textbf{14.5}\% on OpenBookQA, \textbf{14.5}\% on PIQA, and \textbf{6.2}\% on RS, respectively. Hence, MKS2 is capable of markedly improving the overall performance on text-only tasks requiring physical world knowledge. Compared to Vicuna-Llama-2 models with all parameters of Llama-2 updating, our approach stands out by requiring being fine-tuned on only a small fraction of parameters (\textbf{0.25\%} of LLM parameters) while still achieving superior performance on several tasks.

\begin{table*}[t]
\renewcommand\arraystretch{1.10}
  \caption{Zero-shot model performances on natural language processing benchmarks. Models with $^{\dagger}$ indicate that their SFT or RLHF-tuned data are unknown or unused in our work. ``INST-LoRA'' refers to applying the widely-used LoRA technical to fine-tune LLMs with the same text-only instruction data. ``Multimodal-SFT'' represents the multimodal instruction-following data. ``Avg'' refers to the average evaluation score on the total tasks. \textbf{Bold} and \underline{underlined} numbers refer to the best and second-best performance for comparative model variants of Llama-2-7b/13b, respectively.}
  \label{text_results}
  \centering
  \resizebox{\textwidth}{!}{
  \begin{tabular}{l|ccccccc|c}
    \toprule
    Models$\downarrow$ Types $\rightarrow$ &  COQA & StrategyQA & Social IQA & OBQA & PIQA & RS & MMLU & Avg\\
    \hline
    KOSMOS-2 \citep{peng2023kosmos} & -& -& -& -& 72.9& -& -& -\\
    Llama-2-13b-chat$^{\dagger}$~\citep{touvron2023llama} & 37.02 & 37.80 &49.46 &42.89 &67.29 &27.45 & 47.69 & 44.23\\
    Vicuna-Llama-2-13b$^{\dagger}$~\citep{vicuna2023} & 58.21 & 38.82 & 55.85 & 55.46 & 67.01 & 37.02 & \textbf{50.96} & 51.90 \\
    Llama-2-13b-INST-LoRA$_{r=16}$ & 57.68 & 63.73& 63.80 & 58.6 & 71.98 & 38.51  & 46.70 & 57.28\\
    \textbf{MKS2}-Llama-2-13b & \textbf{62.10} & \underline{74.68} & \textbf{65.71} & \textbf{67.6} & \textbf{76.11} & \textbf{41.03} & \underline{48.83} & \textbf{62.30}\\ 
     w/o Multimodal-SFT & 58.77 & \textbf{74.73} & \underline{64.56} & \underline{60.6} & \underline{75.03} & \underline{38.74} & 48.44 & \underline{60.12} \\
     w/o (Multimodal-SFT \& MoMEs) & 54.81 & 68.21 & 62.25 & 54.0 & 67.95 & 35.08 & 46.50 & 55.54\\
     \hline
    Llama-2-7b-chat$^{\dagger}$~\citep{touvron2023llama} & 31.62	& 36.83 & 42.37 &35.3 &64.90	&23.53 & 37.05 & 38.82\\
    Vicuna-Llama-2-7b$^{\dagger}$~\citep{vicuna2023} & 42.58 & 67.58 & 39.71 & 38.2 & 55.62& \underline{29.32} &  \underline{38.94} & 44.56\\
    Llama-2-7b-INST-LoRA$_{r=16}$ & 41.93 & 74.10 & 54.65 & 39.4 &
    53.42 & 27.01 &38.68 & 47.02\\
    \textbf{MKS2}-Llama-2-7b & \textbf{49.38} & \underline{76.15} & \textbf{58.51} & \textbf{54.0} & \textbf{68.19} & \textbf{33.20} & \textbf{39.27} & \textbf{54.10}\\
    w/o Multimodal-SFT & \underline{44.06} & \textbf{76.46} & \underline{57.72} & \underline{50.5} & \underline{67.10} & 28.99 & 37.45 & \underline{51.84}\\
    w/o (Multimodal-SFT \& MoMEs) & 42.84 & 70.46 & 55.42 &37.0
& 60.71 &  25.27 & 37.71 & 47.06\\
    \bottomrule
  \end{tabular}}
\end{table*}

\begin{table*}[t]
\renewcommand\arraystretch{1.10}
  \caption{Model performances on multimodal datasets.``NumImg'' represents the total number of images contained in the pretraining stage. The size of the input image is always 224$^2$ for the following models, and answers are directly generated, except for LLaVA-v1.5 with the image size of 336$^2$. ``$^{\ddag}$'' indicates that the corresponding model employs \textit{the training sets of the following evaluation benchmarks} such as VQAv2, OK-VQA, and OCR-VQA. MKS2-Llama-2-7b$^{\ddag}$ and LLaVA-v1.5 employ the same size of visual instruction tuning data, which is used to compare the performance of introducing the mixture-of-expert architecture.}
  \label{results_vqa}
  \centering
  \resizebox{\textwidth}{!}{
  \begin{tabular}{lc|cccccc|c}
    \toprule
    Models$\downarrow$ Types $\rightarrow$ &  NumImg  & VQAv2 & OK-VQA & STVQA & OCR-VQA& TextVQA & DocVQA & Avg\\
    \hline
    Flamingo~\citep{alayrac2022flamingo} & $>$1B &  49.2 & 41.2 & 19.3 &27.8 &29.0 &5.0 & 28.6\\
    MiniGPT-4 (Vicuna-7b)~\citep{zhu2023minigpt} & 5M  & 44.3 & 32.1 & 14.0 & 11.5 & 18.7 & 3.0 & 20.6\\
    OFA-Large~\citep{wang2022unifying} & 20M  & 40.2 & 19.3& - & -& - &- &- \\
    FROMAGe (OPT-6.7b)~\citep{koh2023grounding} & 3.3M & 44.1 & 20.1 & - & -& - &- &-\\
    mPLUG-Owl$^{\ddag}$~\citep{ye2023mplug} & \textbf{11B} & - & - &29.3 &28.6 &40.3 & 6.9 &- \\
    KOSMOS-2 \citep{peng2023kosmos} & $>$2B & 45.6 & -& - & - & - & - & -\\
    ViperGPT ($>$11B) \citep{suris2023vipergpt} & $>$129M  & - & 48.1 & -& -& -&- & -\\
    ImageBind-LLM (Chinese-LLama-7B) \citep{han2023imagebind} & $>$3M  & - & 51.7 & 15.5 & 23.2 & 24.0&4.0 & -\\
    REVEAL (2B) \citep{hu2023reveal} & $>$12M & & 59.1& - & - & - & - & -\\
    InstructBLIP$^{\ddag}$ (FlanT5{\fontsize{6pt}{13.2pt}\selectfont XL})~\citep{dai2024instructblip} & 129M  & 62.6 & 50.1 & 23.9 & 39.7 & 33.1 & 3.8 &46.0\\
    BLIP-2 (OPT-6.7b)~\citep{li2023blip2} & 129M  & 50.1 & 36.4 & 13.4 &10.6 &21.2 &0.8 & 22.1\\
    BLIP-2 (FlanT5{\fontsize{6pt}{13.2pt}\selectfont XL})~\citep{li2023blip2} & 129M & 42.8 & 25.6 & 15.8 & 26.6& 25.2& 2.9 & 23.2\\
    BLIP-2 (FlanT5{\fontsize{6pt}{13.2pt}\selectfont XXL-11B})~\citep{li2023blip2} & 129M & 45.4 & 27.8 & 21.7 & 30.7 & 32.2 & 4.9 & 27.1 \\
    \hline

    LLaVAR (Vicuna-13b)~\citep{zhang2023llavar} & 1M  & 54.2& 44.9 &
    30.2 & 23.4 & 39.5 & 6.2 & 33.1\\
    \textbf{MKS2}-Llama-2-13b & 2.3M & 54.4 & 45.1  & 28.4 & 35.8 & 37.2 & 6.8 & 34.6\\
    \hdashline[1pt/1pt]
    LLaVA (Vicuna-7b)~\citep{liu2023visual} & 0.6M  & 53.5 & 43.2 &
    22.1 & 11.4 & 28.9 & 4.5 & 27.3\\
    LLaVAR (Vicuna-7b)~\citep{zhang2023llavar} & 1M  & 51.3& 40.6 &
    28.9 & 24.9 & 35.8 & 6.2 & 31.3\\
    \textbf{MKS2}-Llama-2-7b & 2.3M  & 53.3 & 42.1 & 22.3 & 25.2 & 33.1 & 6.7 & 30.5\\
    w/o Text-SFT & 2.3M & 50.2 & 40.8 & 21.5 & 36.5 & 34.2 & 7.4 & 31.7\\
    w/o (Text-SFT \& MoMEs) & 2.3M & 50.1& 41.2 & 21.4 & 35.3 & 34.3 & 7.3 & 31.6\\
    \hdashline[1pt/1pt]
    LLaVA-v1.5$^{\ddag}$ (Vicuna-7b)~\citep{liu2024improved}  & 1M  & 78.4 & 59.8 & 51.3 & 53.1 &	53.1 &  22.4 & 53.0\\
    \textbf{MKS2}-Llama-2-7b$^{\ddag}$ & 1M & \textbf{83.3} & \textbf{64.7} & \textbf{51.6} &  \textbf{54.8} & \textbf{53.2} & \textbf{22.8} & \textbf{55.1}\\  
    \bottomrule
  \end{tabular}}
\end{table*}

\begin{figure*}[t]
    \centering
    \includegraphics[width=0.98\textwidth]{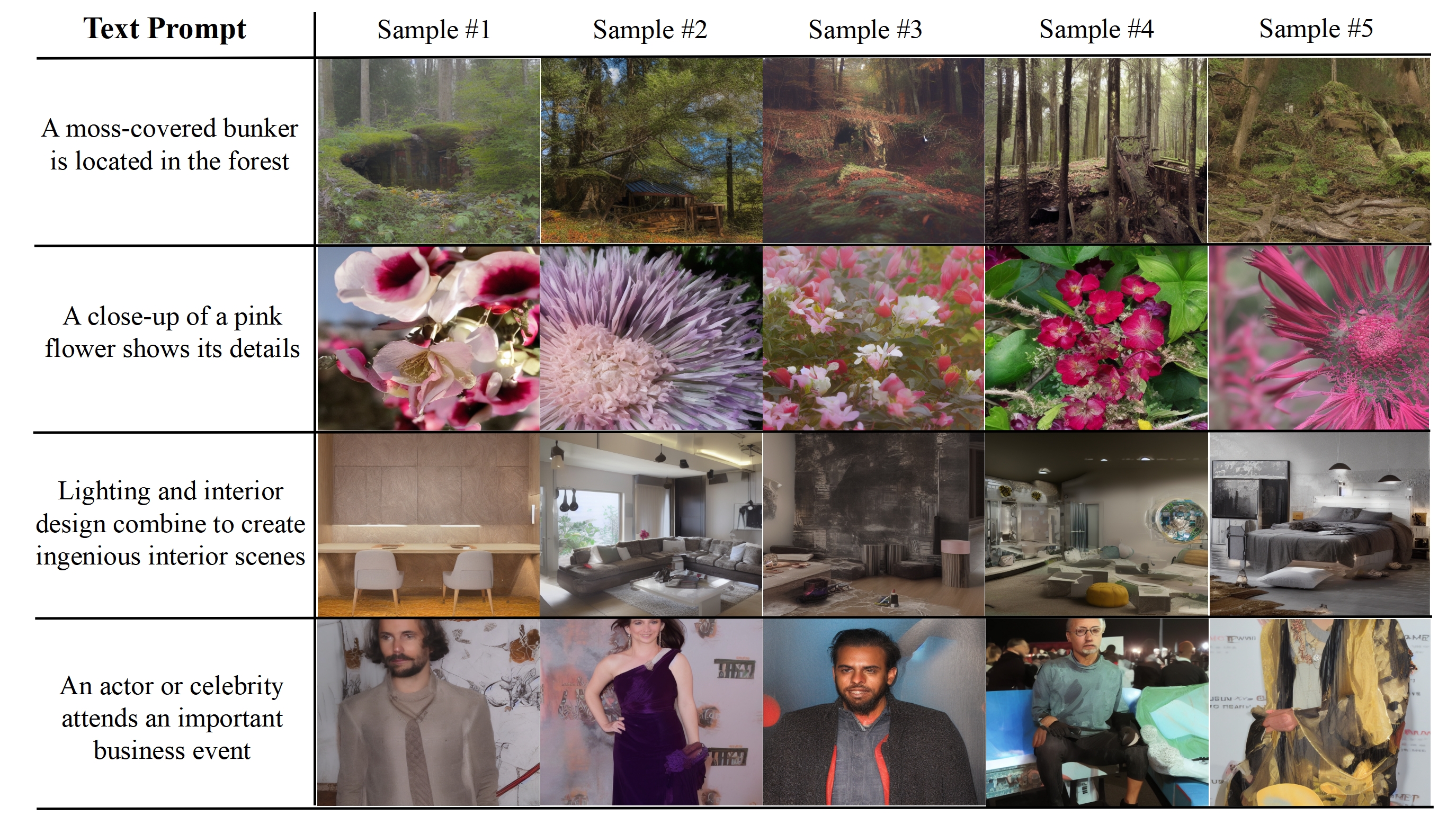}
    \caption{An illustration of cases generated by the pretrained MVM, where we add the generation loss as the supervision object. It's important to highlight that we assess image generation quality primarily to verify whether the added MVM effectively stores text-visual knowledge and establishes the connection between language and vision. Fortunately, the MVM can connect the language to their corresponding imagination.}
    \label{case_generation}
\end{figure*}

\begin{table}[t]
\renewcommand\arraystretch{1.05}
\setlength\tabcolsep{1.3pt}
  \caption{Top@1 accuracy of text-to-image retrieval on Image Retrieval (Flick 30k) and Image captioning Quality. ``$^{\ddag}$'' indicates that the corresponding model is trained on the corresponding training set.}
  \label{ablation_retrieval}
  \centering
  \footnotesize
  \begin{tabular}{l|c|c}
    \toprule
    Models$\downarrow$ Types $\rightarrow$ & CoCo Caption (B@4) &  Image Retrieval (R@1)\\
    \hline
    CLIP~\citep{hafner2021clip} & 38.2 & 68.7\\
    ALIGN~\citep{li2021align} & - & 75.7\\
    FILIP~\citep{yao2021filip} & -& 75.0 \\
    E2E-VLP~\citep{xu2021e2e} & 36.2 & 73.6 \\
    Florence~\citep{yuan2021florence} & - & 76.7 \\
    Oscar~\citep{li2020oscar} & 41.7 & 57.5\\
    CoCa~\citep{yu2022coca} & 40.9 & 76.8\\
    OFA~\citep{wang2022unifying} (20M Images) & 44.9 & 82.2\\
    mPLUG~\citep{ye2023mplug}$^{\ddag}$ & 46.5 & 88.4\\
    GIT~\citep{wang2022git} (0.8B Images) & 44.1 & -\\
    BLIP-2\citep{li2023blip2}$^{\ddag}$ (129M Images) & 42.4 &  89.7\\
    Our Model (2.4M Images) & 40.0 & 80.0\\
    \bottomrule
  \end{tabular}
\end{table}

\textbf{Competitive performance on multimodal benchmarks}. We also present the model’s zero-shot performance on the VQA dataset in Table~\ref{results_vqa}. To gain suitable and robust image embeddings,  we further fine-tuned the visual mapping network for one epoch and froze all other parameters, which does not affect any text-only performance of LLMs. 
Our model demonstrates competitive performance across diverse VQA benchmarks. Comparing MKS2 w/o Text-SFT to the base MLLM (w/o (Text-SFT \& MoMEs) ) reveals no discernible performance degradation on multimodal tasks, indicating that the visual enhancement successfully preserves the model's original textual knowledge. Furthermore, MKS2 outperforms several prior specialized approaches-including those focused on visual feature injection (e.g., ImageBind-LLM), code-based reasoning (e.g., ViperGPT), and multimodal retrieval (e.g., REVEAL). This superior performance validates our novel methodology of using a language-centred learning strategy and a visual memory module to internalize multimodal knowledge directly within the LLM's parameters.
This implies that the addition of visual enhancement in LLMs did not lead to a loss in text-related knowledge while performing LLMs for vision. Moreover, the incorporation of text-only data proves to be a valuable strategy for enhancing the model's proficiency in answering open visual questions. 
While beneficial for QA, mixed instruction data harms scene text recognition, likely due to training data shifts. Optimizing MKS2's performance requires investigating fine-tuning and data distributions.

\begin{table}[tp]
  \caption{Ablation Experiments on five datasets requiring common sense. We explore various data and visual memory sizes to check model performances. All models are built upon Llama-2-7b, where MKS2$^{\clubsuit}$ signifies  MKS2 w/o Multimodal-SFT. "-AM-BM" indicates that it includes an "A" size of visual memory and was trained using "B" number of image-text pairs.}
  \label{ablation_text}
  \centering
  \begin{tabular}{l|cccccc}
    \toprule
    Models$\downarrow$ Types $\rightarrow$ & COQA& OBQA & PIQA & RS & Social IQA \\
    \hline
    Llama-2-7b-chat$^{\dagger}$ & 31.62	 &35.3	&64.90	&23.53 & 42.37\\
    Vicuna-Llama-2-7b$^{\dagger}$ & 42.58	&38.2	&53.42	& 29.32 & 39.71\\
    MKS2$^{\clubsuit}$-410M-2.3M & 44.06 & 50.5 & 67.10 & 28.99 & 57.72\\
    MKS2$^{\clubsuit}$-410M-12.3M & 46.12 & \textbf{50.8} & 64.68\textcolor{blue}{$\downarrow$} & 30.06 & 55.31\textcolor{blue}{$\downarrow$}\\
    MKS2$^{\clubsuit}$-810M-2.3M & 43.78 & 48.6 & 66.97 & 28.89 & 57.21\\
    MKS2$^{\clubsuit}$-810M-12.3M & \textbf{46.56} & 49.8& \textbf{68.10} & \textbf{30.65} & \textbf{57.84}\\
    \bottomrule
  \end{tabular}
\end{table}

\subsection{Visual Storage Analysis}

This section mainly conducts further studies to analyze the effectiveness of modular visual memory (MVM).
we first validate the effectiveness of the MVM by comparing the performance of two configurations: MKS2 w/o (Multimodal-SFT \& MoMEs) and MKS2 w/o Multimodal-SFT, where ``w/o (Multimodal-SFT \& MoMEs) '' shows that we do not utilize the pretrained MVM module. When the MVM is employed, the significant improvement in performance attests to its capacity for external information storage. Given that the MVM is exposed solely to visual information during pretraining, we deduce that it stores visual information effectively.

In Table~\ref{ablation_retrieval}, we can see that our model trained with 2.3M achieves certain image description generation and image retrieval capabilities. This indicates that using a large language model as a text understanding encoder could attain better image-text connection capabilities. This might be because the large language model has larger parameters and stronger language understanding capabilities. However, compared to multimodal encoders pre-trained with tens of millions or hundreds of millions of image-text pairs and supervised fine-tuned multimodal models, there is a large performance gap. Nevertheless, increasing the pre-training data of image-text pairs can make up for this performance gap. Although this is not the focus of this paper, our main objective is to enhance the knowledge scope of the large language model and verify that the proposed MVM is capable of storing visual information.

Additionally, we demonstrated the ability of our model to generate images based on text prompts after the first stage of pre-training in Figure~\ref{case_generation}. From the cases displayed in the figure, we observed that when image generation is added as the loss of the first stage, our model could still adapt and generate relatively good images. In other words, it can imagine relevant visual content based on the text information. The experimental results in Table~\ref{ablation_retrieval} and the image generation quality in Figure~\ref{case_generation} indicate that the introduced language-centric learning method can inspire MVM to store visual information and establish the connection between images and text. By adding a simple MLP to each layer of the pre-trained large language model, we could expand the knowledge storage surface of the large language model, and the relevant parts could be activated through a language-centric learning strategy.

To further examine the MVM’s capability for open-world visual information storage, we varied the number of images in our experiments to assess whether the MVM could accommodate different scales of visual information. This variation directly influences the model’s downstream performance. The experimental outcomes, detailed in Table \ref{ablation_text}, indicate that the downstream performance fluctuates with the size of the training image dataset, under consistent experimental setups. This variation underscores the MVM’s adaptability to different scales of visual information.

\subsection{Ablation Study and Analysis}

\textbf{Effects of MKS2}. 
Comparing the experimental results of MKS2 w/o Multimodal-SFT, MKS2 w/o (Multimodal-SFT \& MoMEs), and Llama-2-7b-INST-LoRA in Table~\ref{text_results}, we observed that the incorporation of visual information into the MVM positively impacted the model's performance on common sense reasoning tasks. This demonstrates that MoMEs can leverage visual knowledge from MVM to improve their understanding and reasoning abilities in various contexts. 
As shown in Table~\ref{text_results}, we performed ablation studies using both the 7B and 13B versions of the Llama model, comparing settings with and without the MVM. For instance, the condition ``without Multimodal-SFT'' was compared to ``without (Multimodal-SFT \& MoMEs)``. In the absence of the MVM module, the model loses its mixture-of-experts capability and reverts to standard fine-tuning, trained solely on plain text with a rank $r=$ 8 configuration, e.g., w/o (Multimodal-SFT \& MoMEs). The comparison between these two variants demonstrates that introducing MVM-which stores visual information as part of a visual expert-leads to noticeable improvements on textual tasks. For example, we observe an average performance gain of 6\% on StrategyQA. This suggests that visual knowledge stored via MVM provides complementary information that aids in textual reasoning.
Additionally, the performances of MKS2 variants w/o Text-SFT and Text-SFT \& MoMEs on multimodal tasks further emphasize its ability to access textual knowledge without significant compromise. The integration of visual information alongside textual data did not hinder the model's capacity to extract and utilize textual knowledge effectively. This suggests that the core text-related capabilities of LLMs remain robust when dealing with multimodal inputs.

\begin{table*}[t]
  \caption{Model performances on \textbf{MMBench}. The language Model and Vision Model represent the backbones of MLLMs. ``TotalParams'' indicates the total parameters of MLLMs. Logical Reasoning (LR), Attribute Reasoning (AR), Relation Reasoning (RR), Fine-grained Perception (Cross Instance) (FP-C), Fine-grained Perception (Single Instance) (FP-S), and Coarse Perception (CP).}
  \label{mmbench}
  \centering
  \resizebox{\textwidth}{!}{
  \begin{tabular}{lccc|ccccccc}
    \hline
    Models$\downarrow$ Types $\rightarrow$ &  Language Model & Vision Model & TotalParams & Overall  & LR & AR & RR& FP-S&FP-C & CP \\
    \hline
    OpenFlamingo~\citep{alayrac2022flamingo} &- & - & 9B & 4.3 &	9.1	&11.4	&3.3	&2.5	&1.6	&1.5\\
    OpenFlamingo v2~\citep{alayrac2022flamingo} & -& - & 9B & 5.7	&11.4&	12.8&	1.4	&5.5	&0.8	&4.0\\
    MiniGPT-4~\citep{zhu2023minigpt} & Vicuna 7B& 	EVA-G 1.1B& 8B & 23.0	&13.6&	32.9&	8.9	&28.8	&11.2	&28.3\\
    MMGPT~\citep{mmgpt} & LLaMA 7B& 	CLIP VIT-L/14 428M & 9B & 16.0	&1.1	&23.8	&20.7&	18.3	&5.2	&18.3 \\
    PandaGPT~\citep{su2023pandagpt}  & Vicuna 13B& ImageBind ViT-H/14 1.2B& 14B & 30.6	&15.3	&41.5	&22.0	&20.3	&20.4	& 47.9 \\
    VisualGLM~\citep{du2022glm} & ChatGLM 6B& 	EVA-G 1.1B	&8B	&	33.5	&11.4	&48.8	&27.7	&35.8	&17.6	&41.5\\
    InstructBLIP~\citep{dai2024instructblip} & Vicuna 7B& 	EVA-G 1.1B	&8B	&	33.9	&21.6	&47.4	&22.5	&33.0	&24.4&	41.1 \\
    LLaVA~\citep{liu2023visual} & LLaMA  7B& 	CLIP VIT-L/14 428M & 8B & 36.2&	15.9&	53.6&	28.6&	41.8&	20.0&	40.4 \\
    LLaMA-Adapter-v2~\citep{gao2023llama} & LLaMA  7B& 	CLIP VIT-L/14 428M	&	8B&	38.9	&7.4	&45.3	&19.2&	45.0	&32.0	&54.0 \\
    G2PT  & LLaMA  7B& 	ViT-G 1.1B	& 8B	&	39.8	&14.8	&46.7&	31.5	&41.8&	34.4	&49.8\\
    mPLUG-Owl~\citep{ye2023mplug}  & LLaMA  7B& 	CLIP VIT-L/14 428M&	8B	&	56.7	&30.7&	65.7	&52.1	&61.0	&45.6&	65.1\\
    Otter-I~\citep{li2023otter} & LLaMA  7B& 	CLIP VIT-L/14 428M	& 9B	&	48.3	&22.2	&63.3&	39.4	&46.8&	36.4	&60.6\\
    Shikra~\citep{chen2023shikra} & Vicuna 7B& 	CLIP VIT-L/14 428M	& 8B	&	60.2&	33.5&	69.6	&53.1	& 61.8	& 50.4	& 71.7\\
    LMEye~\citep{li2023lmeye} & Vicuna 7B & CLIP VIT-L/14 428M	& 8B	& 61.0 & 37.5	& 70.8	& 43.6 & 62.1	& 55.9	& 73.2 \\
    LLaVA-v1.5~\citep{liu2024improved} & Vicuna 7B & CLIP VIT-L/14 428M & 8B& 62.5 & 22.9 & 72.4 & 53.9 & 63.8 & 58.7 & 75.7 \\
    MKS2-Llama-2-7b$^{\ddag}$ ($r=16) $ & Vicuna 7B & CLIP VIT-L/14 428M & 8B & \textbf{73.0} & \textbf{57.1} & \textbf{72.7}   & \textbf{55.6} & \textbf{80.0} & \textbf{75.8} & \textbf{75.7} \\
    \hline
  \end{tabular}}
\end{table*}

\textbf{Impact of visual memory size and data scale}. 
We introduce more image-text pairs from LAION-400M~\citep{schuhmann2021laion} to analyze the impacts of vision memory and data sizes, and the experimental results are shown in Table~\ref{ablation_text}. Our experimental results reveal that enlarging the size of the pre-trained image-text data leads to improvements in the MKS2 model's performance across various downstream tasks. Additionally, we observe that expanding the visual storage module of the MKS2 model not only enhances performance but also contributes to greater stability in this enhancement, as the size of the image-text data increases. These findings suggest a twofold strategy for optimizing the MKS2-based LLMs during the SFT phase: increasing the size of the image-text data and selecting a proportionately larger visual storage size. It is crucial for achieving consistent and stable performance improvements.

\textbf{Could text-only instruction data be effective for enhancing multimodal performance while building MLLMs?}  As discussed in Table~\ref{results_vqa}, text-only instruction data was found to enhance the performance of pre-trained LLMs on various open multimodal problems, particularly open-ended questions that demand a fusion of textual and visual information. However, the introduction of text-only instruction data may lead to a trade-off, as it could potentially diminish the LLMs to answer visual questions involving scene text recognition. These findings underscore the importance of striking a careful balance when incorporating additional data modalities into LLMs, with a keen consideration of task requirements and data characteristics. Such careful integration ensures that overall model performance is optimized without unintended consequences on its ability to handle diverse multimodal challenges.

\textbf{Impact of multimodal instruction data}.
Our experimental results in Table~\ref{text_results} highlight the impact of multimodal instruction data on MKS2's performance. While it effectively enhances the accuracy in addressing questions related to physical world knowledge, it does not provide a substantial boost in solving intricate and complex problems that demand advanced reasoning and strategic thinking. These findings emphasise the importance of tailoring data and approaches to specific task requirements when leveraging multimodal data in LLMs for optimal performance. This may also be because the current visual instruction data is mainly oriented towards visual understanding, and there is less data on visual background knowledge and reasoning. We acknowledge that differences in pre-training data scale and quality may affect model performance. It is worth noting that models such as LLaVA (0.6M images), LLaVA-v1.5 (1M images), and BLIP-2 (123M images) indeed use varying amounts of pre-training data. Nevertheless, LLaVA achieve better performance than BLIP-2 after being fine-tuned on high-quality multimodal instruction data, underscoring the significant role that instruction tuning plays in the downstream task. The primary goal of pre-training in multimodal LLMs is to align visual representations with the semantic space of the language model, thereby enabling better cross-modal understanding. Hence, to ensure a fair comparison with LLaVA-V1.5, we used the same instruction tuning dataset during fine-tuning. This allows us to more directly evaluate the effectiveness of our architectural improvements, rather than reflecting disparities in pre-training data.
In Table~\ref{results_vqa}, our final model MKS2-Llama-2-7b$^{\ddag}$ achieves better performance with the same scale of parameters and data compared to LLaVA-v1.5$^{\ddag}$. It is worth noting that utilizing only a minor segment of visual instruction tuning data in the SFT stage of MKS2 will not effectively enhance the model's multimodal performance.

\begin{table}[t]
\centering
\caption{Performance metrics across LoRA with different ranks ($r=$). FLOPs indicates the forward propagation latency FLOPs per GPU. Params Count refers to the total tuning parameters and the proportion of total model parameters.}
\label{tab:lora_aba}
\resizebox{0.75\linewidth}{!}{
\begin{tabular}{lcccc}
\toprule
Rank ($r=$) & 8 & 16 & 32 & 64 \\
\midrule
FLOPs & 61.0T & 62.8T & 67.0T & 66.3T \\
Training Time & 23h & 23h & 23h & 24h \\
Params Count & 17.8M/0.25\% & 35.7M/0.50\% & 71.5M/1.00\% & 143.3M/2.00\% \\
MMBench & 70.0 & 73.0 & 71.9 & 69.3\\
\bottomrule
\end{tabular}}
\end{table}

\begin{figure*}[t]
    \centering
    \includegraphics[width=0.99\textwidth]{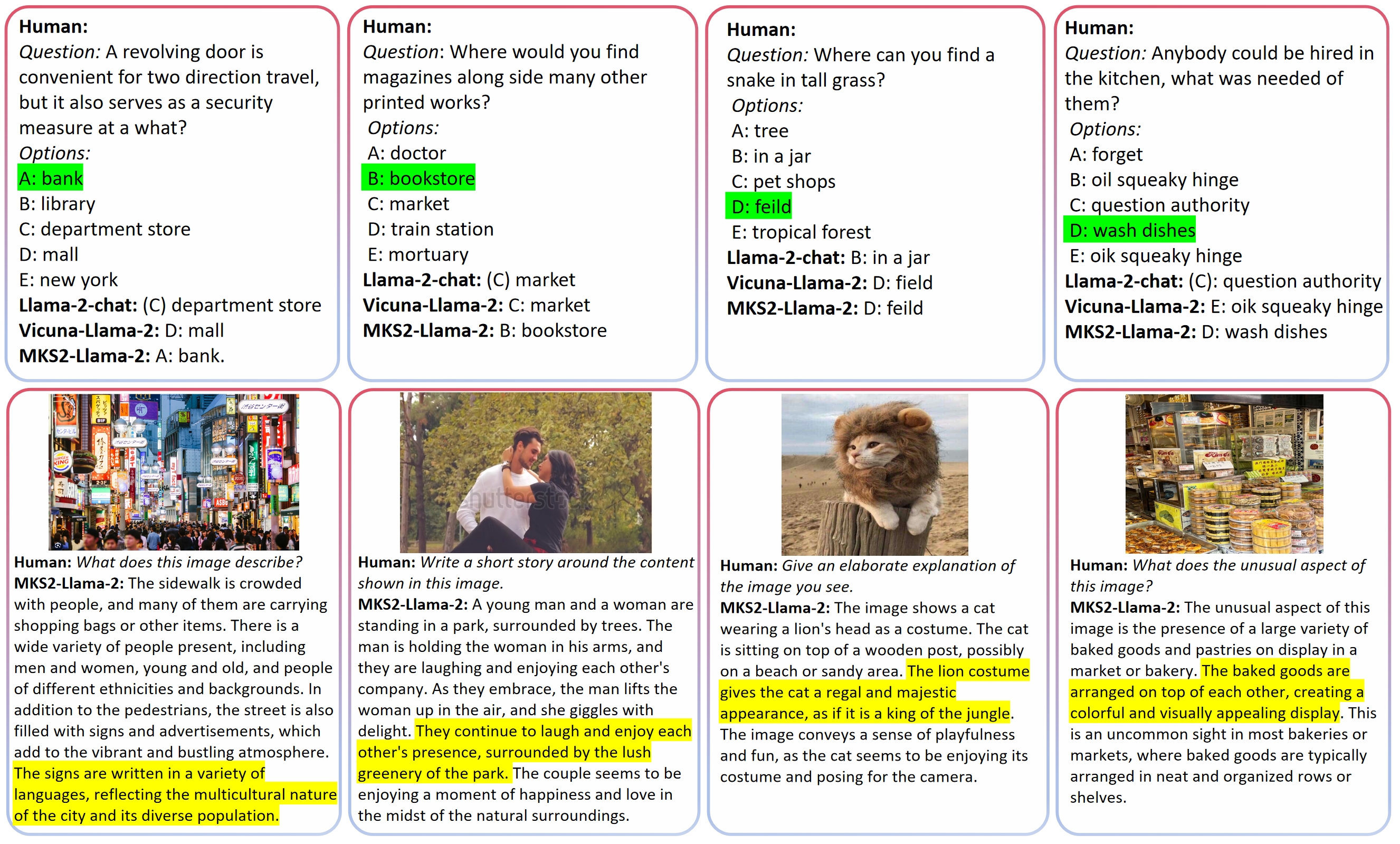}
    \caption{An illustration of cases generated by comparative models and MKS2. \textcolor{green}{Green} words refer to the correct answer in the natural question-answering tasks. \textcolor{yellow}{Yellow} parts represent the interesting and correct content in the response.}
    \label{fig:case_s}
\end{figure*}

\textbf{MoEs for multimodal understanding and reasoning}.
We also explored the MoE architecture-based model, MKS2-Llama-2-7b$^{\ddag}$, focusing on its multimodal understanding and performance. Our model utilizes the same pretraining and visual instruction tuning data as LLaVA-V1.5, with the key distinction being the integration of a visual memory module during the pretraining phase. As shown in Tables \ref{mmbench} and \ref{results_vqa}, our MoE-based multimodal storage and sharing approach outperforms the dense language model alone when constructing a multimodal large model. Additionally, our model did not demonstrate significant improvements with a limited set of visual instruction data in Table \ref{results_vqa}; however, when a larger multimodal dataset LLaVA-665k was used, we observed substantial enhancements in performance on the inference dataset MMBench (Table \ref{mmbench}) and traditional VQA tasks, such as VQAv2 and OK-VQA. This suggests that activating the collaborative capabilities of multimodal experts necessitates more data. The overall training time was about 12 hours on an 8-card A100-80G setup, which may be a worthwhile investment for developing more robust MLLM.



\textbf{Model performance with different LoRA rank ($r$)}.
We conducted an empirical study to investigate the relationship between the number of learnable parameters in LoRA and model performance, which is shown in Table \ref{tab:lora_aba}. Our analysis of LoRA parameters reveals two key findings:
1) Performance Saturates: Model performance on MMBench does not improve significantly with more parameters. 
\begin{wrapfigure}{r}{0.5\textwidth} 
    \centering
    \includegraphics[width=0.45\textwidth]{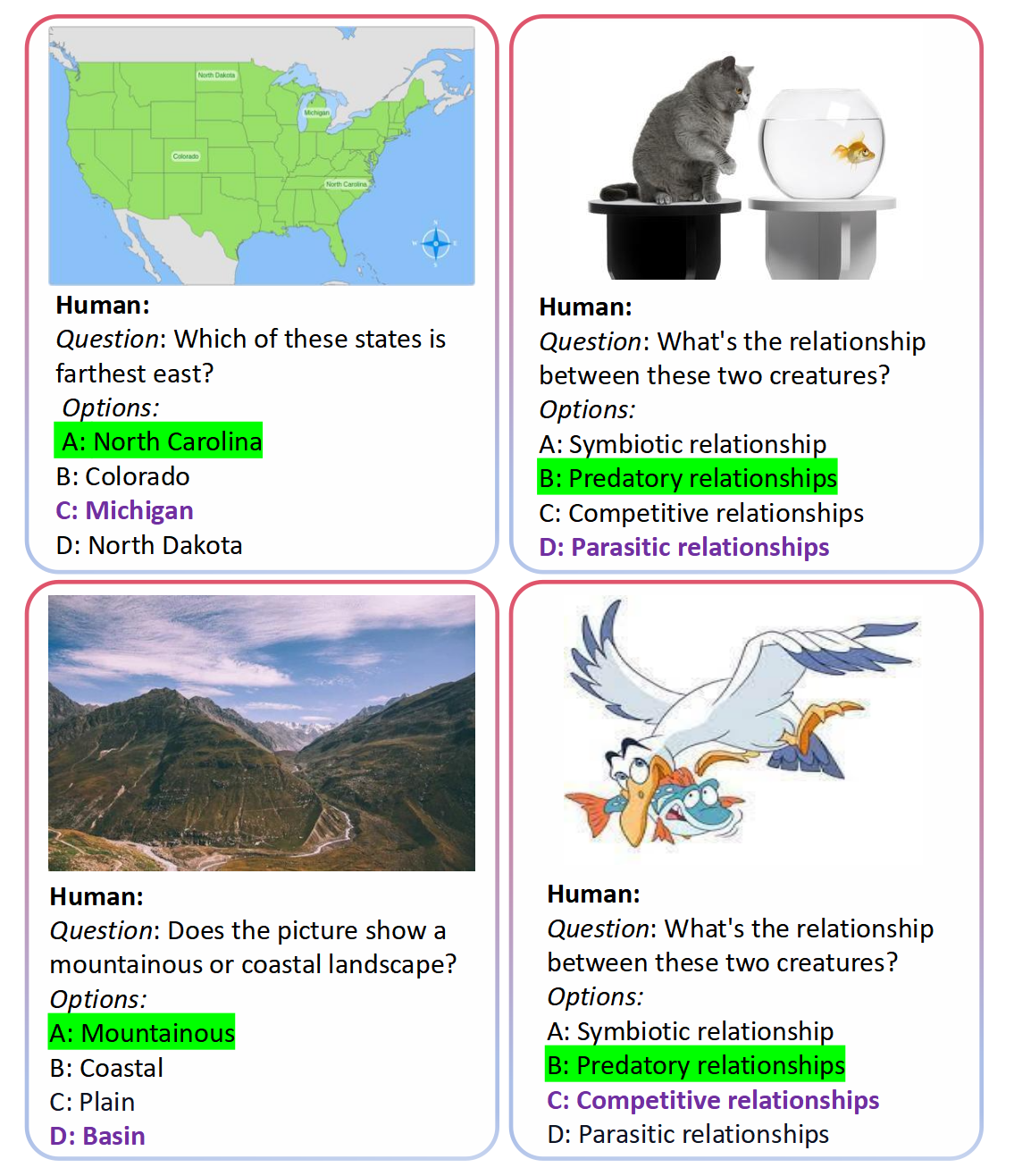}
    \caption{An illustration of cases. Words with a green background refer to the correct answer, yet \textcolor[rgb]{0.5,0,0.5}{Purple} words are results inferred by MKS2.}
    \label{fig:badcase}
\end{wrapfigure}
At a high rank ($r>$16), performance degrades, suggesting excessive parameters lead to overfitting or optimization issues for this task.
2) Efficient Computation: For moderate ranks ($r=$8 to 32), increased parameters did not slow training. 
The added computational load appears to better utilize GPU capacity, maintaining throughput until the parameter count becomes excessive. The overall training time of the model is of the same order of magnitude, and the model performance increase in downstream tasks is not obvious and achieves best on $r=16$. Hence, this suggests that warming up expert activations with limited data does not require a large number of learnable parameters.

\subsection{Case Study}
We present some cases in Figure~\ref{fig:case_s} to further show the overall capability of MKS2-Llama-2 in multimodal understanding and text-only question answering. We can see that the proposed model achieves better performance while answering commonsense Q\&A with physical knowledge, as the above examples shown in Figure~\ref{fig:case_s}. In addition, we also observe that the multimodal understanding capability of MKS2-Llama-2 is also powerful, such as recognising the fun of the cat with a lion's head in the third image.  It could employ the relevant knowledge to enrich the response based on visual clues, e.g., the short story around the content shown in the second image with a young man and a woman. These findings are consistent with the experimental results in the above tables.

However, as shown in Figure \ref{fig:badcase}, we find that our model still struggles with some knowledge-based visual questions, where these questions require the model to combine its visual understanding with external, factual knowledge that is not explicitly contained in the image pixels, and visual-spatial reasoning questions, which demands model to understand and reason about the spatial relationships, orientations, or relative positions of objects within an image. Based on this finding, the model's primary limitations are its deficiencies in knowledge-based visual reasoning and complex visual-spatial reasoning. While it succeeds at straightforward visual recognition (e.g., identifying a coastal landscape), it fails to integrate external world knowledge, as shown by its inability to apply basic geographical concepts to a map. Furthermore, it struggles to interpret dynamic interactions and spatial relationships between objects, erroneously identifying a clear predator-prey scenario as a symbiotic relationship, highlighting a weakness in inferring actions and causal narratives from visual scenes.

\section{Conclusion}
\label{sec:conclusion}

In this paper, we present a new approach named MKS2, that allows LLMs to memorize and employ visual information, achieving multimodal knowledge storage and collaboration in LLMs. MKS2 consists of modular visual memory and a soft mixture of multimodal experts, which are used to store visual information and realise multimodal knowledge collaboration, respectively. To make the visual memory effective, we introduce the language-centric learning framework that includes text-image understanding and generation. We conduct extensive experiments on many natural language processing and visual question-answering tasks. The experimental results show that MKS2 is capable of enhancing the reasoning capability of LLMs and can be used to solve multimodal problems. The ablation study suggests that introducing the MoE architecture in dense pretrained LLMs could improve their language understanding and multimodal performances.

\section{Acknowledgment}

This work is jointly supported by National Natural Science Foundation of China (Grant No. 62422603), Guangdong Basic and Applied Basic Research Foundation (Grant No. 2024B0101050003), and Shenzhen Science and Technology Program (Grant No. ZDSYS20230626091203008).

\section{Contributors}
\label{sec:contributors}

\textbf{Contributors} 

Yunxin Li, Zhenyu Liu, Baotian Hu, Yuxin Ding, and Min Zhang are with the School of Computer Science and Technology, Harbin Institute of Technology, Shenzhen. (e-mail: liyx@hit.edu.cn,  liuzhenyuhit@gmail.com, hubaotian@hit.edu.cn, yxding@hit.edu.cn, and zhangmin2021@hit.edu.cn). 

Wei Wang and Xiaochun Cao are with the School of Cyber Science and Technology, Shenzhen Campus of Sun Yat-sen University. (e-mail: wangwei29@mail.sysu.edu.cn and caoxiaochun@mail.sysu.edu.cn)

\textbf{Corresponding Author}

Baotian Hu

Harbin Institute of Technology, Shenzhen

Email: hubaotian@hit.edu.cn

\bibliographystyle{lychee}
\bibliography{custom}

\end{document}